\documentclass[letterpaper]{article} 
\usepackage[]{aaai2026}  
\usepackage{times}  
\usepackage{helvet}  
\usepackage{courier}  
\usepackage[hyphens]{url}  
\usepackage{graphicx} 
\usepackage{natbib}  
\usepackage{caption} 
\usepackage{algorithm}
\usepackage{algorithmic}
\usepackage{booktabs}
\usepackage{arydshln}
\usepackage{amsmath}
\usepackage{multirow}
\usepackage{multicol}
\usepackage{xcolor}
\usepackage{tablefootnote}
\usepackage{color, colortbl}
\definecolor{Gray}{gray}{0.91}

\usepackage{xcolor}

\usepackage{newfloat}
\usepackage{listings}
\DeclareCaptionStyle{ruled}{labelfont=normalfont,labelsep=colon,strut=off} 
\floatstyle{ruled}
\newfloat{listing}{tb}{lst}{}
\floatname{listing}{Listing}
\graphicspath{figs/}

\newcommand{\judgename}{FineVAU-Judge}
\newcommand{\metricname}{FV-Score}
\newcommand{\benchname}{FineVAU}
\newcommand{\datasetname}{Fine$W^{3}$}

\newcommand{\papertitle}{\benchname: A Novel Human-Aligned Benchmark for Fine-Grained\\Video Anomaly Understanding}
\title{\papertitle}

\author {
    Joao Alexandre Cardeira Pereira\textsuperscript{\rm 1,2,4},
    Vasco Lopes\textsuperscript{\rm 1,2,3},
    João C. Neves\textsuperscript{\rm 1,3},
    David Semedo\textsuperscript{\rm 1,4}
}
\affiliations {
    \textsuperscript{\rm 1}NOVA LINCS, Lisboa\\
    \textsuperscript{\rm 2}DeepNeuronic, Covilhã\\
    \textsuperscript{\rm 3}University of Beira Interior, Covilhã\\
    \textsuperscript{\rm 4}NOVA FCT, Lisboa\\
    jaca.pereira@campus.fct.unl.pt, vasco.lopes@deepneuronic.com, jcneves@ubi.pt, df.semedo@fct.unl.pt
}

\usepackage{bibentry}

\begin{document}

\maketitle

\begin{abstract}
Video Anomaly Understanding (VAU) is a novel task focused on describing unusual occurrences in videos. Despite growing interest, the evaluation of VAU remains an open challenge. Existing benchmarks rely on n-gram-based metrics (e.g., BLEU, ROUGE-L) or LLM-based evaluation. The first fails to capture the rich, free-form, and visually grounded nature of LVLM responses, while the latter focuses on assessing language quality over factual relevance, often resulting in subjective judgments that are misaligned with human perception. In this work, we address this issue by proposing \benchname, a new benchmark for VAU that shifts the focus towards rich, fine-grained and domain-specific understanding of anomalous videos. We formulate VAU as a three-fold problem, with the goal of comprehensively understanding key descriptive elements of anomalies in video: events (\textit{What}), participating entities (\textit{Who}) and location (\textit{Where}). Our benchmark introduces a) \metricname, a novel, human-aligned evaluation metric that assesses the presence of critical visual elements in LVLM answers, providing interpretable, fine-grained feedback; and b) \datasetname, a novel, comprehensive dataset curated through a structured and fully automatic procedure that augments existing human annotations with high quality, fine-grained visual information. Human evaluation reveals that our proposed metric has a superior alignment with human perception of anomalies in comparison to current approaches. Detailed experiments on \benchname~unveil critical limitations in LVLM's ability to perceive anomalous events that require spatial and fine-grained temporal understanding, despite strong performance on coarse grain, static information, and events with strong visual cues.

\end{abstract}

\begin{links}
    \link{Code and Dataset}{https://finevau.github.io}
\end{links}
\begin{figure}[t!]
    \centering
    \includegraphics[width=\columnwidth]{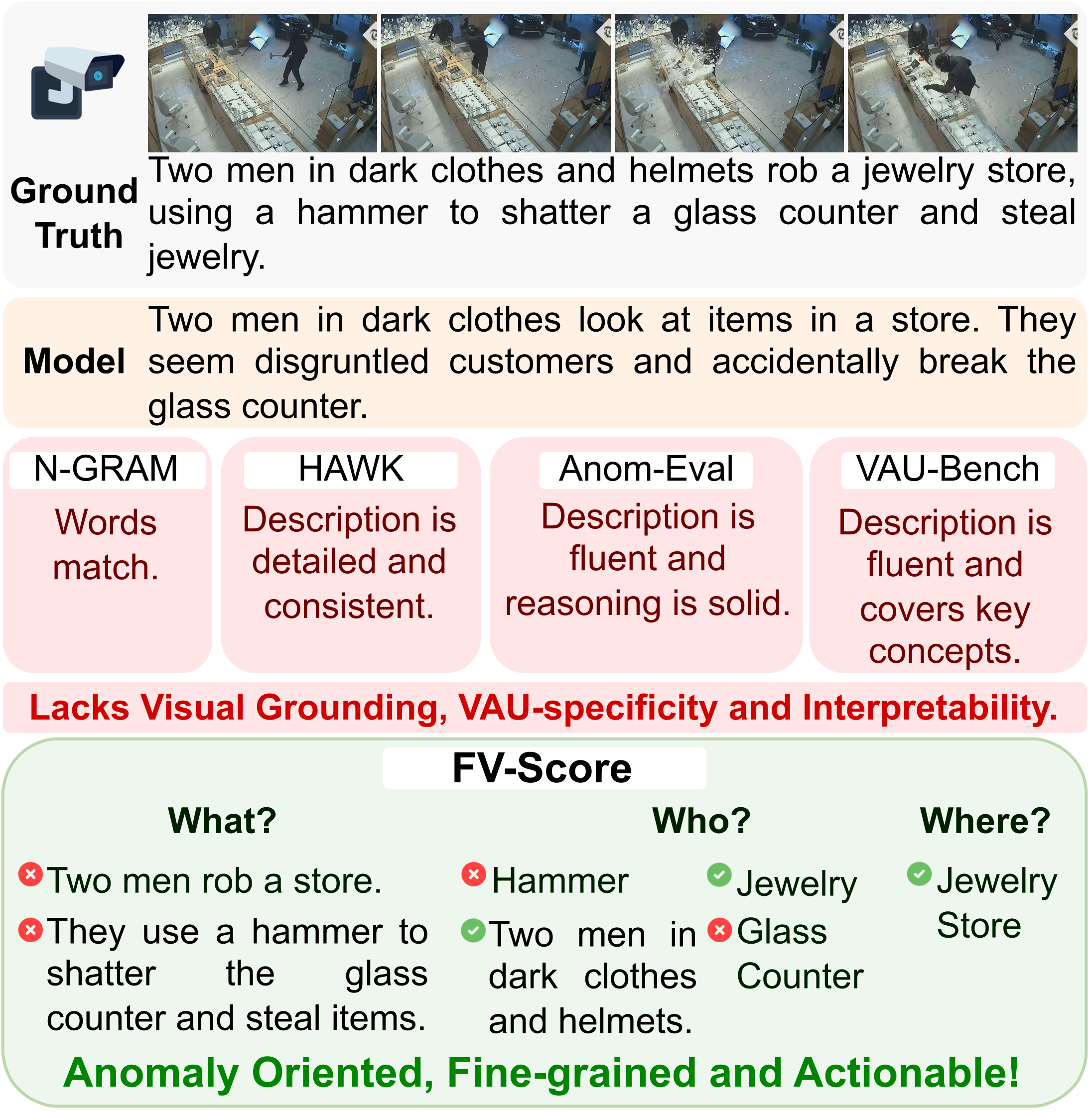}
    \caption{\textbf{Evaluation in Video Anomaly Understanding.} The performance of VAU models is commonly assessed with metrics that a) disregard semantic equivalence; b) focus on language criteria such as fluency and detail; and c) yield non-interpretable, vague scores. Existing metrics fail to signal if the model provides a factually incorrect but coherent description. In contrast, our metric can correctly classify the description as incorrect by decomposing it into the key elements humans rely on to perceive anomalies.}
    \label{fig:teaser}
\end{figure}
\section{Introduction}
The ability to automatically and robustly detect anomalies in video footage has become increasingly critical across a wide range of applications, from public safety to infrastructure monitoring. As video content continues to grow in scale and diversity, there is pressing demand for systems that can robustly process this data and identify unusual or suspicious events without human intervention.
While early anomaly detection systems have shown proficiency in classifying and localizing predefined sets of anomalies~\cite{ref_ucfcrime, ref_varcmp, ref_vera, ref_eventvad}, realistic Video Anomaly Understanding (VAU), which entails a deep understanding of the nuances of abnormal events and the underlying scene, remains an open challenge. The recent emergence of Large Vision-Language Models (LVLMs)~\cite{ref_llava, ref_qwen25vl, ref_llavaov, ref_videollama3, ref_internvl} and their strong generalization capabilities for diverse vision tasks, has inspired significant advancements in VAU, enabling the shift towards more expressive and informative understanding tasks, such as dense video captioning~\cite{ref_uca, ref_holmes, ref_hawk}, video question answering~\cite{ref_surveillancevqa, ref_hawk, ref_vaur1}, and chain-of-thought (CoT) reasoning~\cite{ref_cuva, ref_ecva, ref_vaur1}. 

\begin{table*}[t!]
    \centering
    \resizebox{\linewidth}{!}{%
    \begin{tabular}{llll}
        \toprule
        \textbf{Benchmark} & \textbf{Metric} & \textbf{Criteria} & \textbf{Focus} \\
        \midrule
        UCA~\cite{ref_uca} & N-Gram & Lexical Overlap &  Language \\
        HIVAU~\cite{ref_holmes} & N-Gram & Lexical Overlap &  Language \\
        Hawk~\cite{ref_hawk} &  Both & Lexical\textsection\label{table:fl}, Detail, Reasonability, Consistency &  Language \& Anomaly Description \\
        SurveillanceVQA~\cite{ref_surveillancevqa} & LLM & CI, DO, CU, TU, \textdagger\label{table:ci} &  Language\\
        AnomEval~\cite{ref_cuva} &  LLM & Basic Reasoning, Consistency, Hallucination &  Language \& Anomaly Causality  \\
        VAU-EVAL & LLM  & CA, KC, Fl, In, FC  \textdaggerdbl\label{table:kc}& Language \& Anomaly Reasoning\\
         \rowcolor{Gray}\textbf{\metricname~(Ours) } & \textbf{LLM} & \textbf{Events, Entities, Location} & \textbf{Human-aligned Anomaly Perception}  \\ 
        \bottomrule
    \end{tabular}
    }
    \caption{\textbf{Comparison with SOTA VAU metrics.} Unlike current metrics, which focus on lexical overlap (n-gram based) or on textual fluency and reasoning capabilities (LLM-based), our metric assesses fine-grained understanding of anomaly-specific elements: events (\textbf{\textit{What?}}), entities (\textbf{\textit{Who?}}) and location (\textbf{\textit{Where?}}). The following abbreviations are used: \textsection~Lexical Overlap; \textdagger~CI: Contextual Integration; DO: Detail Orientation; CU: Contextual Understanding; TU: Temporal Understanding; \textdaggerdbl~CA: Classification Accuracy; KC: Key Concept Alignment; Fl: Linguistic Fluency; In: Informativeness; FC: Factual Consistency.
    }
    \label{tab:sota}
\end{table*}


Despite rapid progress, current VAU reference benchmarks largely overlook evaluation, adopting inadequate metrics, weakly correlated with human judgments, hindering the accurate assessment of the true capabilities of proposed models. These can be split into two categories:
1) traditional n-gram based metrics (e.g., BLEU, ROUGE-L)~\cite{ref_bleu, ref_rouge, ref_meteor, ref_cider}, which measure lexical overlap rather than factual accuracy or contextual understanding, and thus are inherently ill suited for free-form outputs of modern LVLMs; and 2) LLM-based metrics~\cite{ref_cuva, ref_ecva, ref_hawk, ref_vaur1}, often directly adopted from general video understanding tasks~\cite{ref_vcgpt}, which focus on textual fluency and  overall coherent reasoning capabilities, lacking the necessary granularity to pinpoint VAU-specific aspects, resulting in subjective scores that are misaligned with human perception of anomalies.


To address this pressing gap in VAU evaluation, we propose \benchname, a novel automatic and highly human correlated benchmark, that drives the focus towards a rich, fine-grained and domain-specific understanding of anomalies in videos, covering key aspects of human anomaly perception. Namely, by identifying the key structural characteristics of a video anomaly, we formulate \benchname~ as a three-perspective problem, comprising comprehensive understanding of 1) events (\textbf{\textit{What?}}), 2) entities (\textbf{\textit{Who?}}), and 3) location information (\textbf{\textit{Where?}}) from anomaly videos (see Figure~\ref{fig:teaser}). Grasping these perspectives is key to enable effective and coherent model reasoning about the existence of an anomaly in video.

We enable our structured evaluation through a novel dataset, curated with a fully automatic LVLM-assisted pipeline that systematically decomposes and structures existing human-labeled anomaly description annotations into high-quality knowledge, carefully determining the \textit{What, Who, Where} dimensions.
Leveraging this anomaly structuring, we propose \metricname, a novel LLM-based metric that frames evaluation as a multi-part detection problem, with the goal of extracting the \textit{What, Who and Where}, from LVLMs' reasoning and generated responses. 
In this setting, \metricname~brings three key properties: a) it breaks anomaly video evaluation in individual dimensions, providing finer-grained, structured and explainable signals, b) achieves strong correlation with human annotations, and c) pushes LVLMs to identify anomalies in video by reasoning and grounding responses across these three dimensions.

Our experiments underscore the advantages of \metricname's nuanced and fine-grained feedback, unveiling that state-of-the-art LVLMs struggle to report and perceive anomalous events that lack strong visual cues, despite a more accurate understanding of static entities and scene elements. Consequently, our benchmark represents a new, challenging frontier for human-aligned VAU.

In summary, our main contributions are as follows:
\begin{itemize}
\item \textbf{\benchname~}, a novel benchmark for Video Anomaly Understanding (VAU) that emphasizes fine-grained, human-aligned evaluation grounded in the core components of anomaly comprehension: events (\textit{What}), entities (\textit{Who}), and location (\textit{Where}).

\item \textbf{\metricname}, an LLM-based metric that performs key element detection on LVLM answers, providing interpretable and actionable feedback that is tightly aligned with human perception

\item \textbf{\datasetname}, a high-quality dataset that enriches existing high quality anomaly video annotations with \textit{What, Who, Where} information through a systematic and scalable augmentation pipeline leveraging LVLMs.

\item Extensive experiments across diverse LVLMs demonstrate the importance of our evaluation, revealing critical blind spots in current models’ ability to capture complex and subtle anomalies.

\end{itemize}


\section{Related Work}
\label{sec:related_work}
\subsubsection{Video Anomaly Detection.} Early works primarily focus on Video Anomaly Detection (VAD) by framing it as a video-level classification problem
(e.g., Shoplifting, Robbery) or localizing abnormal frames \cite{ref_ucfcrime, ref_xdviolence}. Despite providing coarse-grain anomaly signals, 
these methods offer a broad, high level understanding of the video, limiting actionable insights into the nature or context of the anomaly. We address the more challenging task of VAU, requiring fine-grained understanding about the core elements of anomaly videos.

\subsubsection{Video Anomaly Understanding.} More recently, the advent of Large Vision-Language Models (LVLMs) \cite{ref_llavavideo, ref_qwen25vl, ref_internvl, ref_videollama3} has allowed for rapid progress towards VAU. The pioneering work in UCA~\cite{ref_uca} introduces dense human-labeled captions to describe the events of videos in the popular UCF-Crime~\cite{ref_ucfcrime} dataset. HAWK~\cite{ref_hawk} proposes sets of synthetically generated video descriptions and question-answer pairs. Similarly, Holmes-VAU~\cite{ref_holmes} proposes anomaly video descriptions at clip, event and video-level. More recently, ECVA~\cite{ref_ecva} (originally CUVA~\cite{ref_cuva}) introduces manually-annotated anomaly reasoning data, covering the cause, description and result of abnormal events. VAU-Bench~\cite{ref_vaur1} proposes synthetic Chain-of-Thought (CoT) reasoning for anomaly, explaining events by analyzing causal factors, temporal dynamics, and contextual cues. These works 
lack rich object and scene dimension information, which are tightly coupled with the nature of the anomaly, and rely on synthetic annotations containing hallucinations. 
We leverage high-quality human-labeled annotations and augment them with rich, verifiable information covering three fundamental anomaly understanding dimensions.


\subsubsection{VAU Evaluation.} Current VAU evaluation methods, largely borrowed or adapted from general-purpose scenarios, and suffer from critical limitations. N-gram-based metrics (e.g., BLEU~\cite{ref_bleu}, METEOR~\cite{ref_meteor}, ROUGE-L~\cite{ref_rouge}, CIDEr~\cite{ref_cider}), used in UCA, Holmes-VAU and HAWK, measure direct lexical overlap between a reference and a predicted caption, thus failing to accurately capture the inherent intricacies of anomalies in free-form responses and reasoning traces, and penalizing factually correct but lexically divergent answers. 
Addressing these, LLM-based judges have been proposed.
AnomEVAL \cite{ref_ecva}, adopted in the ECVA benchmark, focuses on assessing causal reasoning by evaluating a model's ability to understand the cause and result of anomalies. 
However, the scores provided lack fine-grained grounding to anomaly characteristics. SurveillanceVQA-589K (Surv-VQA) \cite{ref_surveillancevqa} employs a multi-dimensional evaluation protocol to assess multiple criteria, but relies on  subjective judgments of correctness across broad categories, rather than a verifiable detection of specific visual elements crucial for VAU. 
Finally, VAU-Eval \cite{ref_vaur1}, introduced in VAU-Bench, assesses a model's reasoning capabilities against structured question-answering and rationale annotations. Its focus on language-specific aspects and holistic criteria are insufficiently granular to determine accurate perception of anomaly-specific information. We address this by proposing \judgename, covering fine-grained anomaly dimensions: \textit{What, Who, Where}.



\section{Fine-Grained Video Anomaly Understanding}
\label{sec:evaluation}


With \benchname, we formulate VAU as the goal of comprehensively understanding the key structure of anomaly videos according to human perception of anomalies, grounded in a hierarchy composed by three main structural dimensions: events (\textit{What}), involved entities and their attributes (\textit{Who}) and location (\textit{Where}).

\subsection{Problem Formulation} 
Let $V = \{f_1, f_2, \dots, f_T\}$ be an untrimmed video, represented as a sequence of $T$ frames. The goal of a VAU model $M$ is to generate a natural language report of the video $R = M(V)$.  We define a structured ground truth, $G$, for each video $V$. This ground truth is a set of fundamental anomaly elements, partitioned according to the three hierarchical dimensions:
\begin{enumerate}
    \item \textbf{What (Events):} $G_{\text{what}} = \{e_1, e_2, \dots, e_{N_e}\}$, a set of $N_e$ textual descriptions capturing the key actions (e.g., "sets fire"), interactions (e.g., "fighting") and isolated state changes (e.g., "explosion") occurring in the video.
    \item \textbf{Who (Entities):} $G_{\text{who}} = \{w_1, w_2, \dots, w_{N_w}\}$, a set of $N_w$ concise textual descriptions of the involved actors or objects, including their salient visual attributes (e.g., clothing, color, age group).
    \item \textbf{Where (Location):} $G_{\text{where}} = \{l_1, l_2, \dots, l_{N_l}\}$, a set of $N_l$ attributes detailing the scene where the events unfold.
\end{enumerate}
The complete ground truth of a video is the union of these sets: $G = G_{\text{what}} \cup G_{\text{who}} \cup G_{\text{where}}$. The quality of a model's report $R$ is measured by its coverage of the critical elements of the ground truth $G$. To measure this coverage, we replace set hard membership in $G$ by a semantic-aware membership function $m_{\theta}(g,G_{*})$, with parameters $\theta$, $g$ being a ground truth element, and $G_{*}$ the reference set. Then, we define a structural scoring function,
\begin{equation}
\mathcal{J}_{\text{dim}}(R) = \sum_{g_i \in G_{\text{dim}}}  m_{\theta}(g_i, R),
\label{eq:scoring}
\end{equation}
where $dim \in \{what,who,where\}$, which scores the presence and correct mention of each ground truth element $g_i \in G_{\text{dim}}$ in the generated report $R$, and calculates the total score. 
Unlike previous works, which encompass complex scales and broad criteria, we argue for straightforward scoring instructions, simplifying the task for an LLM judge and increasing interpretability. Thus, we use a binary scale for \textit{Who} and \textit{Where}, and a ternary scale for the \textit{What} dimensions:
\begin{center}
\setlength{\tabcolsep}{3pt}
\renewcommand{\arraystretch}{1.1}
\fbox{%
\begin{tabular}{cl|cl}
\multicolumn{2}{c|}{\textbf{Binary (\textit{Who}, \textit{Where})}} & \multicolumn{2}{c}{\textbf{Ternary (\textit{What})}} \\
\hline
0 & $\leftarrow$ Missing / incorrect & 0 & $\leftarrow$ Missing / incorrect \\
1 & $\leftarrow$ Present, correct & 0.5 & $\leftarrow$ Partial, minor errors \\
  &                               & 1 & $\leftarrow$ Accurate, complete \\
\end{tabular}%
}
\end{center}
The ternary scale provides flexibility to deal with scenarios where the answer partially covers the elements of the ground truth (e.g., $R$ contains \textit{"Two people are having an heated discussion."} and $g$ is \textit{"Two men are fighting."}). The membership function $m_{\theta}$ strictly follows the scale in the aforementioned table.

The overall quality of the report $R$ is then quantified by a scoring function $\mathcal{S}(R)$, which separately aggregates scores across the three dimensions of ground truth elements:
\begin{equation}
\begin{split}
\mathcal{S}(R) ={} \lambda_{\text{what}} \cdot&\mathcal{J}_{\text{what}}(R)  + \lambda_{\text{who}} \cdot \mathcal{J}_{\text{who}}(R) \\
& + \lambda_{\text{where}}  \cdot \mathcal{J}_{\text{where}}(R)
\end{split}
\label{eq:score}
\end{equation}
where $\lambda_{\text{what}}, \lambda_{\text{who}}, \lambda_{\text{where}}$ are weights controlling the relative importance of each dimension. \benchname~defines VAU as the goal of generating a natural language report $R$, for a given video $V$, that maximizes the score $\mathcal{S}(R)$ with respect to its ground truth $G$.

\subsection{\metricname~and \judgename}
We define $\mathcal{S}(R)$ as \textbf{\metricname~} and $\mathcal{J}$ as \textbf{\judgename}, a LLM judge which given an a dimension ground truth set $G_{\text{dim}}$, and the report $R$ for a video, attests semantic membership $m_{\theta}$ for each ground truth element $g\in G$.  

Following MovieChat and VideoChatGPT, two well-established video understanding benchmarks, we materialize the membership function $m_{\theta}$ using a frontier LLM model, Gemini-2.5-Flash~\cite{ref_gemini} that allows judgments using a triplet of \{$R$, $G$, $P$\}, where $P$ is a highly detailed prompt that instructs the model to provide a structured output with the membership scores for each ground truth element $g\in G$, according to a LVLM model output report $R$, and the membership scale defined in the previous section. We detail $P$ in the supplementary material. 


\section{\textit{What, Who, Where} Dataset}
\label{sec:dataset}

We build the ground truth $G$ of our \benchname~benchmark by curating \textit{\textbf{W}hat, \textbf{W}ho, \textbf{W}here} (\datasetname), a novel dataset containing fine-grained and structured information from anomaly videos.  

\subsection{Structured Annotation Scheme}

The core novelty of our dataset is a comprehensive and structured annotation scheme that maps the human perception of anomaly videos, according to the three dimensions defined in \benchname.

At the \textit{What} dimension, the dataset captures key actions, interactions and isolated state changes in the video as a chain of discrete, atomic or highly correlated events. 
The \textit{Who} dimension, provides detailed information regarding the entities involved in the events. Each entity is assigned a unique \textit{identifier} that concisely describes it (e.g., "assailant"). Concurrently, event descriptions (e.g., "The man holds a gun...") are explicitly linked to the unique identifiers of all the entities involved. Additionally, each entity contains information about its \textit{category} (e.g., "person", "vehicle"), and a rich set of observable, category-specific \textit{visual attributes}: \textit{clothing}, approximate \textit{age group}, \textit{gender} and \textit{distinguishing feature} (e.g., "a large beard") for a person; \textit{size}, \textit{color} and \textit{brand} for a vehicle; and at least \textit{color} and \textit{size} for a general object. Finally, at the \textit{Where} dimension, the dataset provides information regarding the location where events take place, including the physical \textit{environment} (e.g., "jewelry store"), \textit{time of day} (e.g., night time), \textit{lighting conditions}, \textit{crowd density} and a \textit{salient feature} that uniquely identifies the scene (e.g., "a large painting on the wall"). 

\begin{figure}[t]
\centering
\includegraphics[width=\columnwidth]{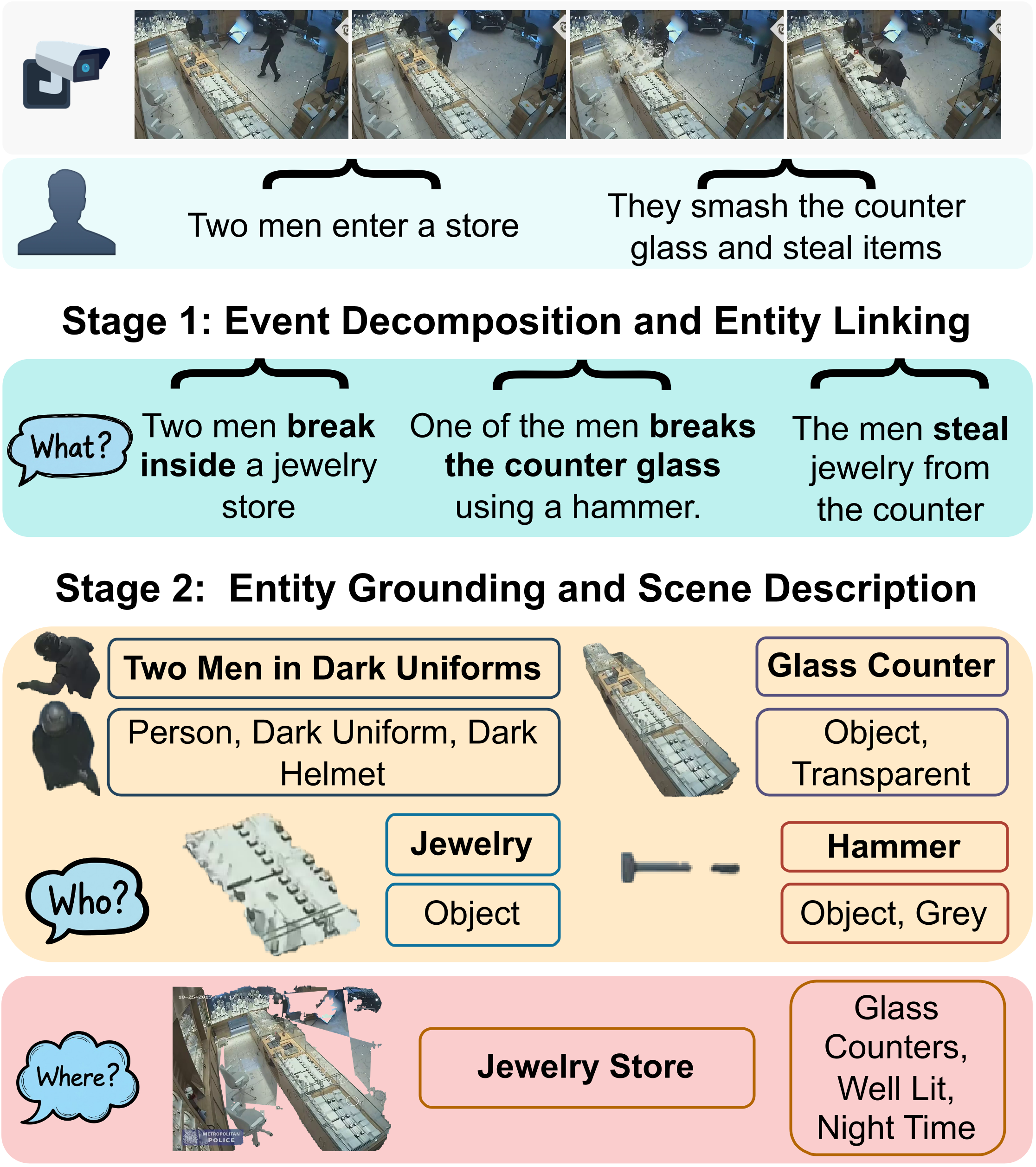} 
\caption{\textbf{Our two-stage, fully automated pipeline for scalable annotation of fine-grained VAU data.} We ground our annotation process on high-quality human annotations, and use a LVLM to: 1) augment and refine existing events and identify the entities involved; an 2) augment entity and location information with fine-grained physical attributes.}
\label{fig:annotation_process}
\end{figure}

\subsection{Annotation Pipeline}
Our dataset requires a high level of granularity in information that is not present in current VAU datasets. Therefore, we develop a fully automated and scalable pipeline to enrich and augment existing, human-labeled VAU data with high quality, fine-grained and structured anomaly-oriented information. We split the annotation process into two stages, as depicted in Figure ~\ref{fig:annotation_process}, and leverage assistance from a LVLM. We employ Gemini-2.5-Pro~\cite{ref_gemini} due to its
strong multimodal understanding and long context capability. For each video, we uniformly sample frames at 1 fps and provide them alongside the original UCA annotations to the model.

\subsubsection{Stage 1: Event Decomposition and Entity Linking.} The first stage of our pipeline augments and refines existing event annotations from UCA~\cite{ref_uca}. An LVLM processes raw, human-generated event descriptions, together with their respective videos, to 1) decompose complex sentences into a chain of fine-grained, causally linked atomic events; 2) complement annotations by identifying unmentioned events or objects that are clearly observable in the video; and 3) identify all participating entities for each event and assign concise identifiers, which are then explicitly linked to the respective event.

\subsubsection{Stage 2: Entity Grounding and Scene Description.} This stage builds on the output of the first stage. We leverage the LVLM to 1) augment linked entities with rich and fine-grained information regarding their observable physical attributes; and 2) identify and describe the physical properties of the location where the events take place. Further information regarding the prompts for the LVLM and resulting annotations can be seen in the supplementary material.

\subsection{Dataset Statistics} 
\begin{figure*}[!t]
    \centering
    \begin{minipage}[t]{0.31\textwidth}
        \centering
        \includegraphics[width=\linewidth, height=4cm, keepaspectratio]{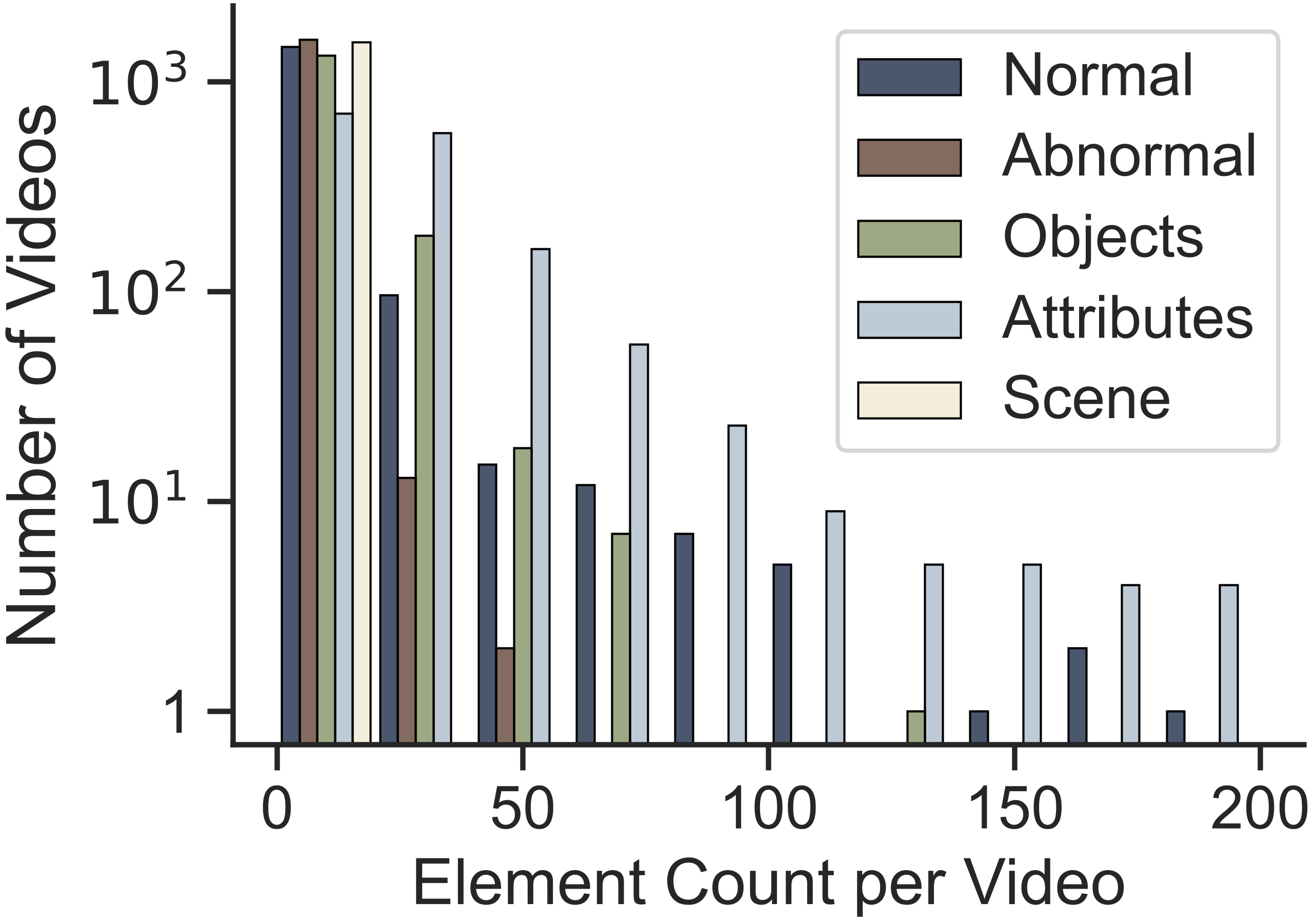}
        \caption{\textbf{Annotation count per dimension.} We split events into normal and abnormal, and separate counts of entities and their physical attributes. The large number of annotations stems from the granularity of our annotations.}
        \label{fig:hist_elements}
    \end{minipage}
    \hfill
    \begin{minipage}[t]{0.31\textwidth}
        \centering
        \includegraphics[width=\linewidth, height=4cm, keepaspectratio]{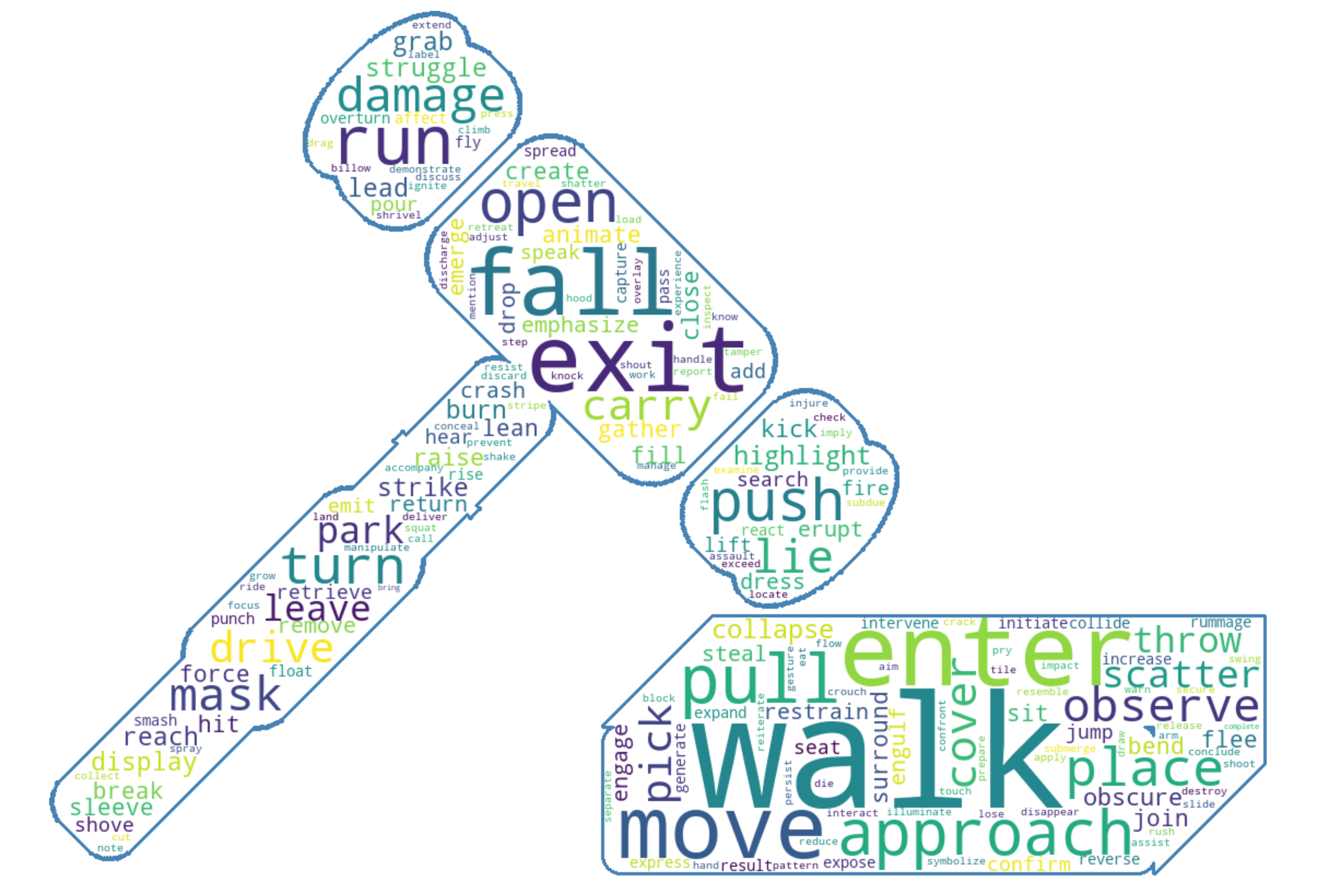}
        \caption{\textbf{Word Cloud for event annotations in \datasetname.} Events often describe motion (e.g., "run", "walk"), interaction ("exit", "approach") and abnormality ("fall", "damage", "struggle", "grab", "kick")}
        \label{fig:wordcloud}
    \end{minipage}
    \hfill
    \begin{minipage}[t]{0.31\textwidth}
        \centering
        \includegraphics[width=\linewidth, height=4cm, keepaspectratio]{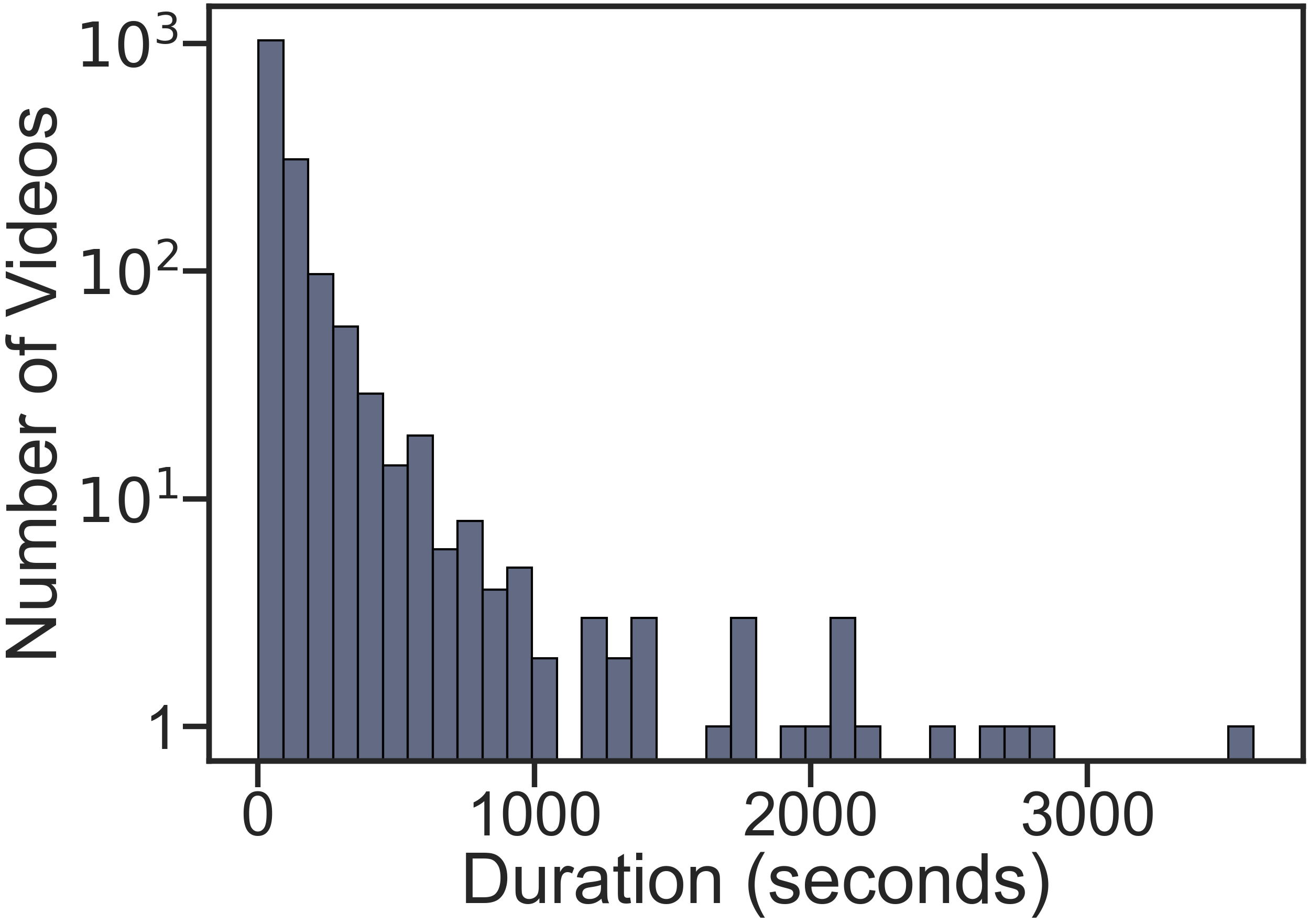}
        \caption{\textbf{Histogram of Video Duration.} Our dataset contains challenging long videos with up to 1h duration. This is expected given the source of the videos of our dataset, composed of CCTV footage.}
        \label{fig:hist_videos}
    \end{minipage}
\end{figure*}

Our dataset contains a total of 1544 videos. These videos contain a total of 17813 events, from which 13393 are normal and 4420 are abnormal. These events are associated with a total of 59392 entities, which in turn reference a total of 74593 individual attributes. Finally, there are 7669 annotated location attributes. Figure \ref{fig:hist_elements} shows the distribution of the number of annotations per video, for all annotation dimensions. At the event dimension, we distinguish abnormal and normal events. As expected, there is a much larger number of normal events per video, since abnormal events usually occur infrequently and in small temporal windows. Figure~\ref{fig:wordcloud} plots the word cloud of event annotations, showing a clear trend towards anomaly and movement related topics. Figure \ref{fig:hist_videos} shows the distribution of the duration of videos in our dataset, originally sourced from CCTV footage, often containing several minutes or even hours of video. There is also a broad range of possible locations, including public streets, highways, shopping centers or private households. Depending on the location and the time of day, there may be large amounts of entities performing a wide variety of different actions (e.g., people walking in a crowded street, customers shopping in a convenience store with several items). The combination of these factors, coupled with our fine-grained approach to VAU, results in a densely annotated, extremely challenging benchmark that enables the true assessment of the capability of LVLMs to fully grasp the complexity of anomalies, establishing a new frontier for VAU.    


\begin{table}[t!]
\centering
\resizebox{\columnwidth}{!}{%
\begin{tabular}{@{}lcccc@{}}
\toprule
\textbf{Metric} & \textbf{PCC $\rho \uparrow$} & \textbf{1-R$^2 \downarrow$} & \textbf{Kd $\tau \uparrow$} & \textbf{Sp $\tau \uparrow$} \\
\midrule
\multicolumn{5}{l}{\underline{\textit{N-gram Baselines}}} \\
CIDEr~\cite{ref_cider}  & -0.63 & 0.60 & -0.59 & -0.58 \\ 
BLEU\textdagger~\cite{ref_bleu} & 0.19 & 0.96 & 0.17 & 0.17 \\ 
METEOR~\cite{ref_meteor} & 0.45 & 0.80 & 0.41 & 0.40 \\ 
ROUGE-L~\cite{ref_rouge} & 0.47 & 0.78 & 0.43 & 0.44 \\ 
\midrule
\multicolumn{5}{l}{\underline{\textit{LLM Judge Baselines}}} \\
AnomEVAL~\cite{ref_ecva} & 0.42 & 0.82 & 0.39 & 0.37 \\
VAU-EVAL~\cite{ref_vaur1} & 0.53 & 0.72 & 0.49 & 0.47 \\
\midrule
\textbf{\metricname} & \textbf{0.61} & \textbf{0.63} & \textbf{0.56} & \textbf{0.56}\\
\bottomrule
\end{tabular}%
}
\caption{\textbf{Correlation of VAU metrics with human judgment.} \textdagger~BLEU is the mean score of BLEU-\{1 to 4\}.}
\label{tab:human_eval}
\end{table}

\section{Assessing \metricname's Human Correlation}
We conduct a comprehensive human evaluation study to validate the alignment of \metricname~with human perception of anomaly report quality. Our study utilizes 60 videos randomly sampled from the UCF-Crime dataset \cite{ref_ucfcrime}. We recruit 8 human experts with computer vision or multimodal evaluation background and assign each video to 3 different experts, resulting in 180 total ranking judgments. For each video, experts rank three reports of varying quality generated by Gemini 2.5-flash \cite{ref_gemini}: 1) a high-quality report covering all critical information; 2) a medium-quality report omitting some key details; and 3) a low-quality report failing to describe the anomalies. To mitigate bias, reports are presented in randomized order without disclosing their quality level. The study achieves a Pairwise Percentage Agreement score of 68\%, indicating substantial inter-annotator agreement, particularly given the subjective nature of the task.

We measure the correlation between state-of-the-art VAU evaluation metrics and human rankings by employing four standard agreement measures~\cite{ref_capture}. We use Pearson Correlation Coefficient (PCC $\rho$) to assess the linear relationship between metric scores and human scores, and $1-R^{2}$ to measure the unexplained variance, where lower values signify a better fit. To evaluate ordinal association, we use both Kendall’s Tau (Kd $\tau$) and Spearman’s Rho (Sp $\tau$), which measure the agreement in the rankings produced by the metric and by human judges. Results are available in Table \ref{tab:human_eval}. Our study does not consider the metric used in SurveillanceVQA~\cite{ref_surveillancevqa} nor HAWK~\cite{ref_hawk}, as they are strictly defined for QA tasks.

Evaluation reveals that \metricname~achieves a superior alignment with human judgment over all baselines. \metricname~achieves the highest scores across all correlation measures (with exception to CIDEr in unexplained variance), achieving a Pearson correlation of \textbf{0.61} and a Kendall's Tau of \textbf{0.56}. This performance marks a clear improvement over the strongest n-gram baseline, ROUGE-L~\cite{ref_rouge} (PCC $\rho$ - 0.47, Kd $\tau$ 0.43). Critically, BLEU~\cite{ref_bleu} and CIDEr~\cite{ref_cider} show systematic disagreement with humans. Surprisingly, LLM-based metrics perform similarly to their n-gram counterparts, despite utilizing strong GPT-based judges. These findings support the hypothesis that the structured and fine-grained evaluation provided by our metric is better aligned with human perception of anomalies.   

\begin{table}[!t]
    \centering
    \resizebox{\columnwidth}{!}{%
    \begin{tabular}{ccccccc}
    \toprule
         $\mathbf{\lambda_{\text{what}}}$ & $\mathbf{\lambda_{\text{who}}}$ & $\mathbf{\lambda_{\text{where}}}$ & \textbf{PCC $\rho \uparrow$} & \textbf{1-R$^2 \downarrow$} & \textbf{Kd $\tau \uparrow$} & \textbf{Sp $\tau \uparrow$} \\
         \midrule
          1.0 & 1.0 & 1.0 & 0.51 & 0.74 & 0.46 & 0.47  \\
          1.0 & 1.0 & 2.0 & 0.47 & 0.77 & 0.42 & 0.42  \\
          2.0 & 1.0 & 1.0 & 0.56 & 0.69 & 0.50 & 0.50  \\
          1.0 & 2.0 & 1.0 &\textbf{0.61} & \textbf{0.63} & \textbf{0.56} & \textbf{0.56}  \\
         \bottomrule
    \end{tabular}
    }
    \caption{\textbf{Ablations on \metricname~weights} reveal that a strong weight for entity components achieves a higher correlation with human judgment, indicating that humans value an accurate perception of involved entities.}
    \label{tab:ablations_lambdas}
\end{table}

\begin{table*}[!t]
\centering
\resizebox{\textwidth}{!}{%
\begin{tabular}{@{}l ccccc ccccc ccc@{}}
\toprule
& \multicolumn{5}{c}{\textbf{Overall Performance}} & \multicolumn{5}{c}{\textbf{Location}} & \multicolumn{3}{c}{\textbf{Entity}} \\
\cmidrule(lr){2-6} \cmidrule(lr){7-11} \cmidrule(lr){12-14}
\textbf{Model} & \textbf{Location} & \textbf{Event} & \textbf{Entity} & \textbf{Attribute} & \textbf{All} & \textbf{Lighting} & \textbf{Env} & \textbf{Crowd} & \textbf{Time} & \textbf{Salient} & \textbf{Person} & \textbf{Vehicle} & \textbf{Others} \\
\midrule
VideoLLaMA3 & 40.3 & 6.5 & 24.3 & 10.2 & 19.3 & 44.1 & 64.7 & 30.0 & 35.4 & 27.4 & 20.8 & 22.2 & 27.5 \\
LLaVA-OV    & 58.3 & 13.0 & 41.1 & 19.9 & 32.2 & 65.1 & 80.1 & 42.2 & 60.0 & 44.1 & 38.5 & 37.6 & 44.0 \\
Qwen2.5-VL  & 70.8 & 9.1 & 38.3 & 20.3 & 32.9 & 80.2 & 83.6 & \textbf{68.0} & \textbf{80.7} & 41.8 & 29.6 & 37.9 & 44.5 \\
LLaVA-VID   & 65.7 & 14.4 & 44.0 & 21.0 & 35.0 & 65.1 & \textbf{87.0} & 56.8 & 69.0 & 50.8 & 42.2 & 38.0 & 47.4 \\
InternVL3   & \textbf{71.8} & \textbf{18.0} & \textbf{51.2} & \textbf{25.5} & \textbf{40.5} & \textbf{80.4} & 86.6 & 59.3 & 79.7 & \textbf{53.1} & \textbf{54.0} & \textbf{44.8} & \textbf{51.5} \\
\midrule
Mean        & 61.3 & 12.2 & 39.8 & 19.4 & 32.0 & 67.0 & 80.4 & 51.3 & 65.0 & 43.4 & 37.0 & 36.1 & 43.0 \\
\bottomrule
\end{tabular}%
}
\caption{\textbf{Results of SOTA LVLMs on \benchname} reveal clear difficulties in our benchmark. Scores are computed using $\lambda_{\text{what}}=1.0$, $\lambda_{\text{who}}=2.0$, $\lambda_{\text{where}}=1.0$ based on our human correlation study (Table~\ref{tab:ablations_lambdas}).
}

\label{tab:results}
\end{table*}

\subsubsection{Ablations on $\lambda_{\text{what}}$, $\lambda_{\text{who}}$ and $\lambda_{\text{where}}$.} We conduct ablations on the weights used for the scores of different dimensions in \metricname~(see Table ~\ref{tab:ablations_lambdas}). While current metrics often focus mostly on event understanding (\textit{What}), our experiments demonstrate that humans highly value reports that correctly identify the main entities involved and accurately describe them (\textit{Who}). This finding further supports the validity of our fine-grained and structured approach to VAU.

\section{Experiments and Results}

\subsection{Experimental Setup}
We leverage \benchname~to evaluate five state-of-the-art open-source LVLMs, namely Qwen2.5-VL-7B~\cite{ref_qwen25vl}, InternVL3-9B~\cite{ref_internvl}, VideoLLaMA3-7B~\cite{ref_videollama3}, LLaVA-Video-7B~\cite{ref_llavavideo} and LLaVA-OneVision-7B~\cite{ref_llavaov}. To facilitate reproducibility, we adopt the widely used lmms-eval~\cite{ref_lmms_eval} platform. All our experiments are done in a zero-shot setting, using the original model weights, and a model temperature of 0. All prompts used are provided in the supplementary material. We process videos following each model's default frame sampling strategy as implemented in lmms-eval.

\begin{figure*}[t!]
\centering
\includegraphics[width=\linewidth]{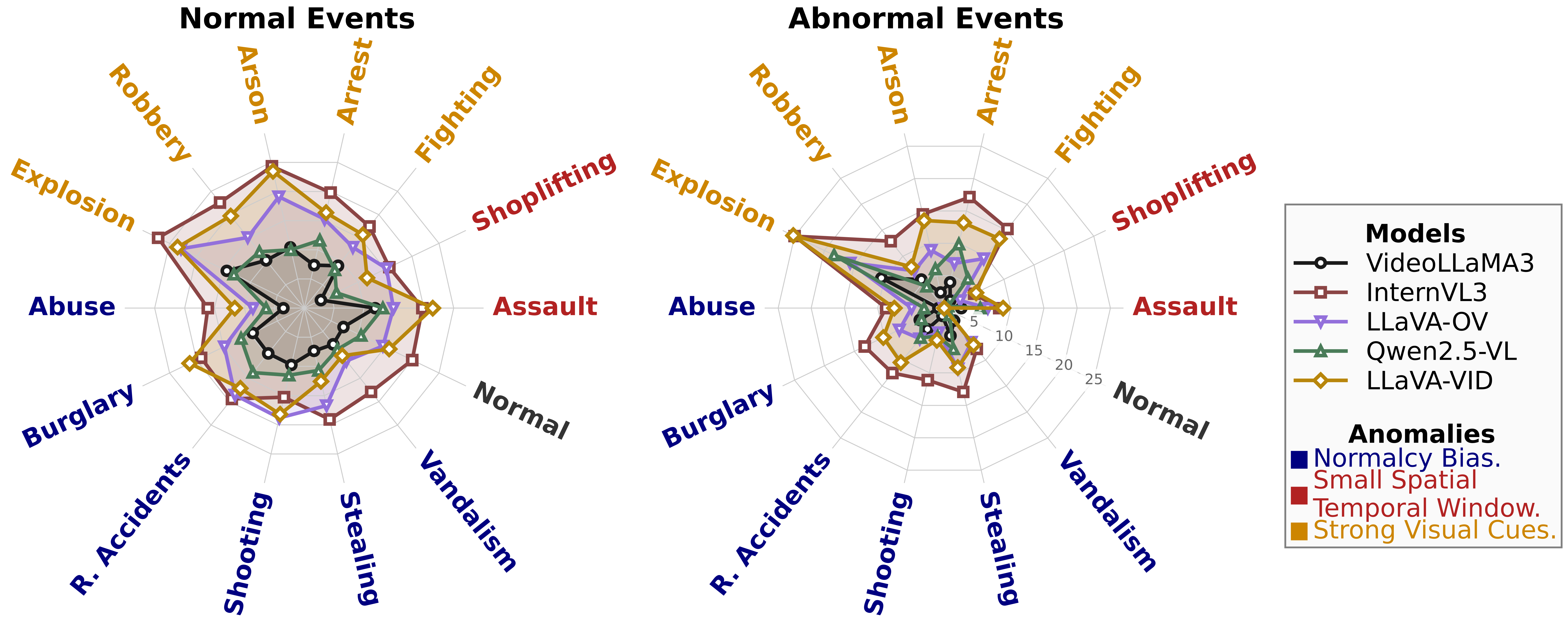} 
\caption{\textbf{LVLM's performance breakdown at event dimension.} Performance per high level anomaly category, summarizing the nature of the events depicted in the video (e.g., \textit{Fighting} contains events depicting the conflict escalation from peaceful coexistence to a physical altercation). More details on expected events per category are provided in the supplementary material.
}
\label{fig:events_plot}
\end{figure*}

\subsection{Results}
Table \ref{tab:results} presents the performance of five LVLMs on \benchname, revealing several important takeaways. 

\subsubsection{LVLMs are stronger at perceiving static and coarse grain information.} LVLMs show significantly better performance at reporting location information, with a mean accuracy of $61.3\%$. We hypothesize that the strong image understanding pretraining of LVLMs (which comprise an image understanding vision encoder) results in a strong capability at grounding static and coarse grain information. This hypothesis is further sustained by their performance at Entity dimension. Despite a low  Entity Mean performance of $39.8\%$, it still surpasses Event and Attribute dimensions. A closer look at performance for individual location attributes evidences once again a similar pattern, since models excel at identifying the physical environment of the videos, and are also capable of perceiving lighting conditions, time of day, and crowd density. At the entity dimension, comprehensive identification of vehicles and people is surprisingly tougher for LVLMs in comparison to other categories. This is likely due to the fact that people and vehicles are the most common objects and are usually more predominant, making them more challenging to report in detail, in contrast with other single unit (e.g., buildings, weapons) and low frequency (e.g., ATMs, animals) categories.

\subsubsection{LVLMs struggle with spatial and temporal fine-grained understanding.}
Unlike for static and coarse grain information, LVLMs struggle significantly on fine-grained understanding, both on the spatial and temporal axis. This is evidenced by the low accuracy on individual object attributes, contrasting with coarse grain object identification. We argue that this difficulty is grounded on the training bias of LVLMs, which is largely composed of general-purpose videos with high resolution, high quality and less clutter, in contrast with the low resolution and low quality of available anomaly videos. Nonetheless, the major struggle of LVLMs lies on identifying all events in anomaly videos, with models exhibiting a mean accuracy of merely 12.2\%. Figure \ref{fig:events_plot} compares the accuracy of LVLMs at the event dimension, according to the high level video category, for both normal (leftmost plot) and abnormal (rightmost plot) events. Noticeably, results once again corroborate our hypothesis that LVLMs perform better in events when strong visual cues are available: \textit{Explosions} and \textit{Arson}, commonly accompanied by flashes of bright light, fire and debris; \textit{Arrests}, which usually involve police officers in characteristic uniforms and vehicles; and \textit{Fights}, which frequently cause high commotion for surrounding individuals. However, anomalies that occur in smaller spatial and temporal windows and require a higher understanding of visual elements and behaviors, are much more challenging. This is the case with \textit{Shoplifting}, which requires understanding sudden behaviors such as placing small items in a bag and leaving without paying.

\subsubsection{LVLMs are biased towards normalcy.} Another noticeable pattern observable in Figure~\ref{fig:events_plot}  is that despite globally achieving low performance, LVLMs are still more capable at understanding \textit{Normal} events, even in videos that contain abnormalities. We argue that LVLMs are biased towards normalcy, and therefore frequently conflate abnormal events for normal ones (e.g., a fight is depicted as a conversation). This high degree of hallucination is not seen inversely, since LVLMs less frequently hallucinate abnormal events in normal situations, as evidenced by their superior performance in the latter. We hypothesize that the low performance of LVLMs in normal events emerges instead from their inability to recall the high number of events annotated in our dataset. Examples of common LVLM hallucinations and failures to recall granular information can be seen in the supplementary material.

\subsubsection{InternVL3 achieves the top performance across all dimensions.} Noticeably, despite lower context sizes, smaller pretraining corpus and lower scene performance, LLaVA-OneVision~\cite{ref_llavaov} and LLaVA-VID~\cite{ref_llavavideo} are more capable of understanding events in comparison to large context and corpus alternatives (Qwen-2.5-VL and VideoLLaMA3). This gap is an additional proof of the critical disconnect between understanding the static context of a video and the anomalous events within it.

\section{Conclusions and Future Directions}

In this work, we introduce \benchname{}, a novel benchmark that addresses the critical gap in Video Anomaly Understanding (VAU) evaluation by shifting the focus to a fine-grained and structured assessment of LVLM comprehension across events (\textit{What}), entities (\textit{Who}), and location (\textit{Where}), which are key aspects in human perception of anomalies. Through our proposed \metricname, supported by an LLM-based~\judgename, and the \datasetname~dataset, we conduct an extensive evaluation of five LVLMs and unveil a critical weakness:
\textbf{While these models are capable of perceiving static scenes and entities, they fundamentally fail to comprehend the fine-grained attribute details and subtle events that occur in small spatial and temporal windows, often hallucinating normalcy}. 
This crucial finding, made possible by our anomaly structuring and \benchname, evidences clear next steps towards developing targeted training to mitigate hallucinations and induce detailed, factual understanding, using our structured data. Pairing such data with rigorous benchmarks such as ours is essential for training and validating the next generation of models capable of truly robust video anomaly understanding.


\clearpage
\section{Acknowledgments}
This work is supported by NOVA LINCS (UID/04516/2025) with the financial support of FCT.IP; and Fundação para a Ciência e Tecnologia ref. 2023.03647.BDANA.

\bibliography{aaai2026}

\appendix
\twocolumn[{%
  \centering
  {\Large\bfseries \papertitle\par}
  \vspace{0.6em}
  {\large Supplementary Material\par}
  \vspace{1.5em}
}]

\begin{figure*}
    \centering
    \includegraphics[width=\linewidth]{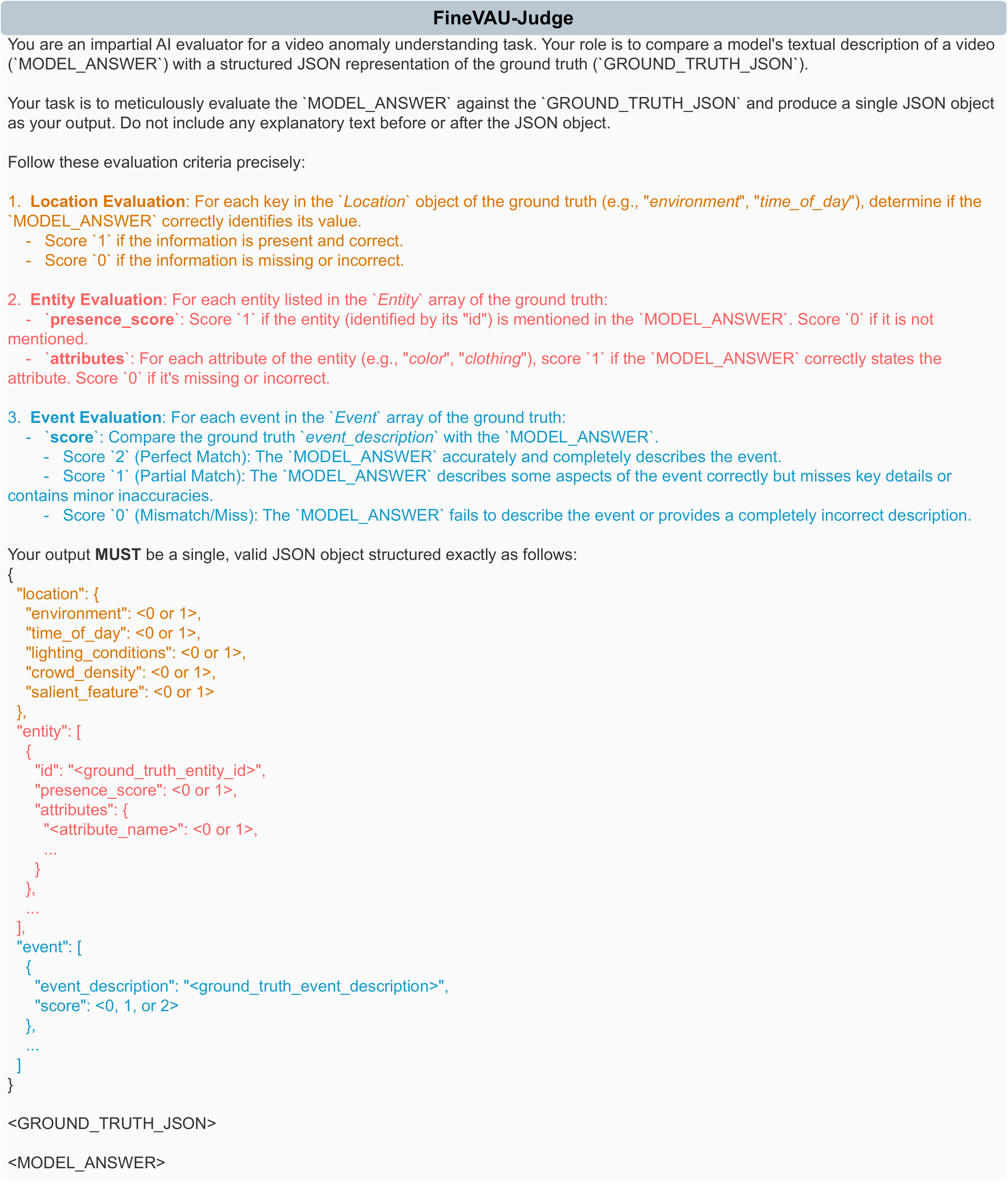}
    \caption{\textbf{\judgename~prompt.} Prompt used to elicit a structured output with the membership scores for each ground truth element from \judgename. We use Gemini-2.5-flash~\cite{ref_gemini} as the judge.}
    \label{fig:prompt_judge}
\end{figure*}

\begin{figure*}
    \centering
    \includegraphics[width=\linewidth]{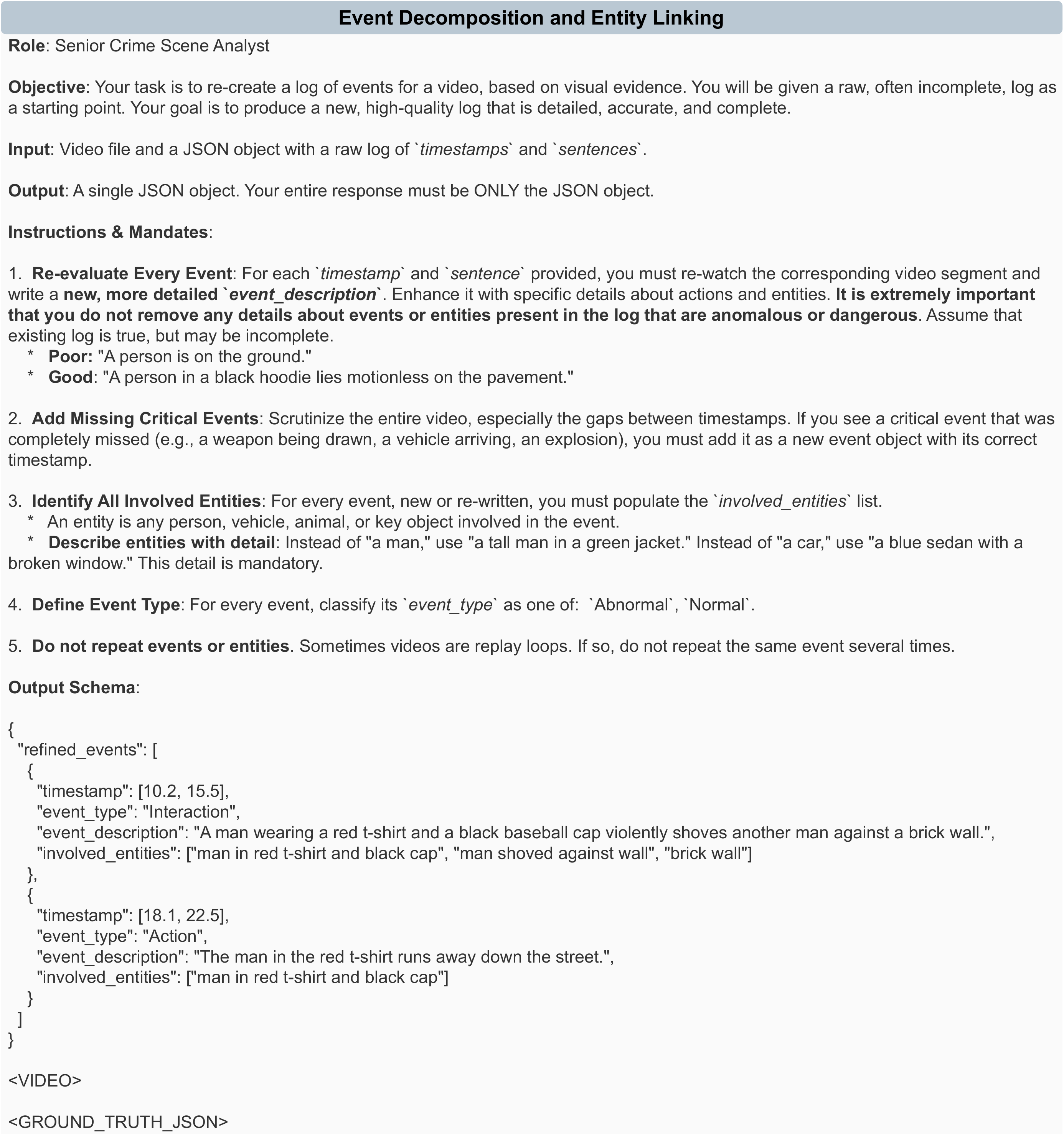}
    \caption{\textbf{Stage 1 Prompt for Dataset Annotation.} Prompt used to augment and refine existing event annotations. We use Gemini-2.5-Flash~\cite{ref_gemini} as the LVLM annotator.}
    \label{fig:prompt_annotation_stage1}
\end{figure*}

\begin{figure*}
    \centering
    \includegraphics[width=\linewidth]{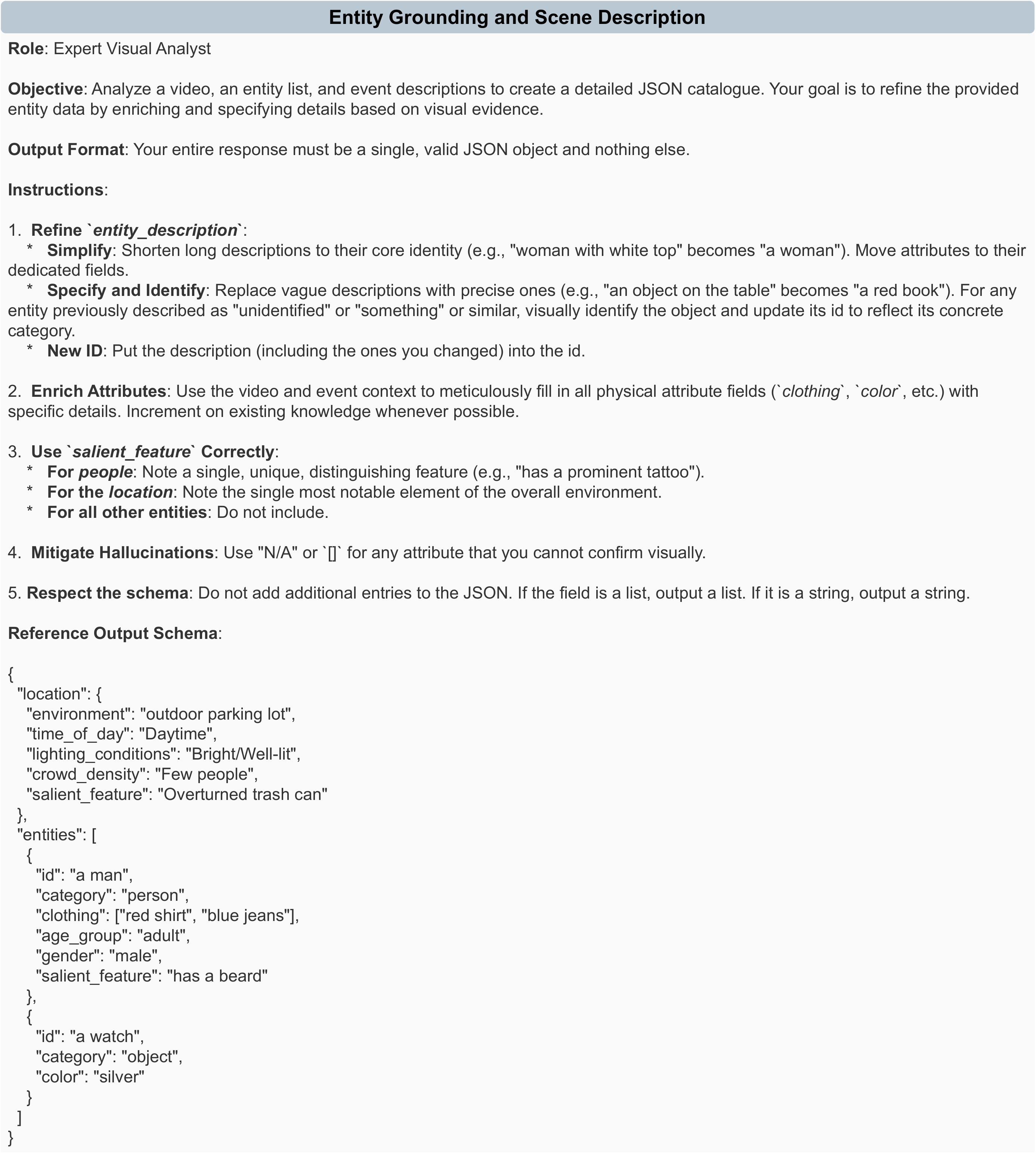}
    \caption{\textbf{Stage 2 Prompt for Dataset Annotation.} Prompt used to augment entity and location attributes. We use Gemini-2.5-Flash~\cite{ref_gemini} as the LVLM annotator.}
    \label{fig:prompt_annotation_stage2}
\end{figure*}

\begin{figure*}
    \centering
    \includegraphics[width=\linewidth]{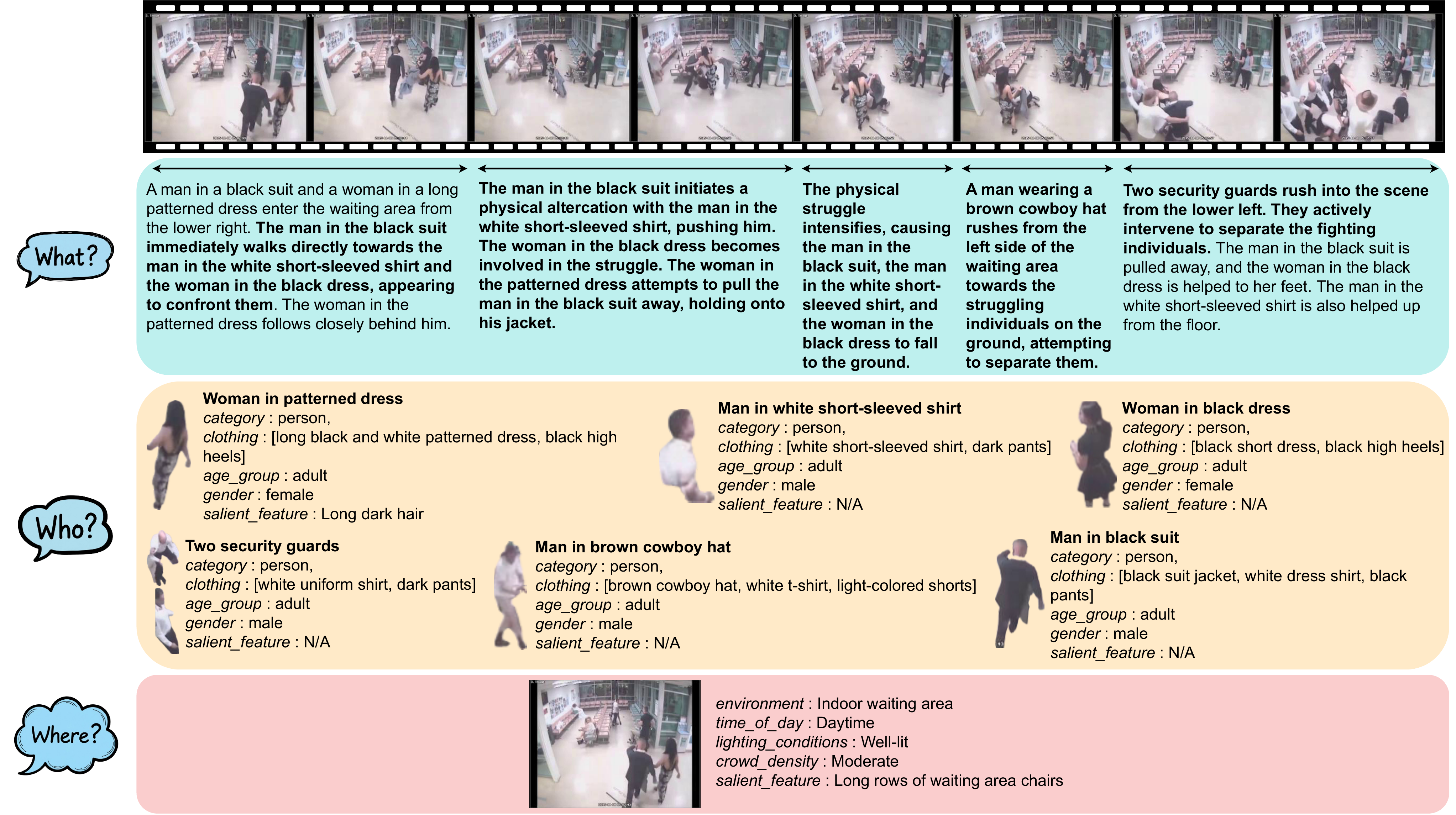}
    \caption{\textbf{Annotation example from \datasetname.} Our dataset contains highly detailed and fine-grained information regarding events (\textit{What}), their involved entities and their physical attributes (\textit{Who}), and the location where events happen, as well as physical attributes of this location (\textit{Where}). To mitigate hallucinations, common in synthetic VAU data, we ground events in preexisting data from an high quality human labeled source (UCA~\cite{ref_uca}), and augment event, entity and location data by instructing the LVLM annotator to only annotate information that is clearly visible in the video. For information that may not clearly be visible, we instruct the LVLM to annotate "N/A" for "Not Applicable". In the example illustrated in the image, the model can't observe distinguishing features for some of the entities annotated. This may indicate: 1) the distinguishing features are already present elsewhere in the data (e.g., in "man in brown cowboy hat", the brown cowboy hat is already in the clothing) or they are simply not clearly visible (e.g., "woman in black dress").} 
    \label{fig:annotation_example}
\end{figure*}

\begin{figure*}
    \centering
    \includegraphics[width=\linewidth]{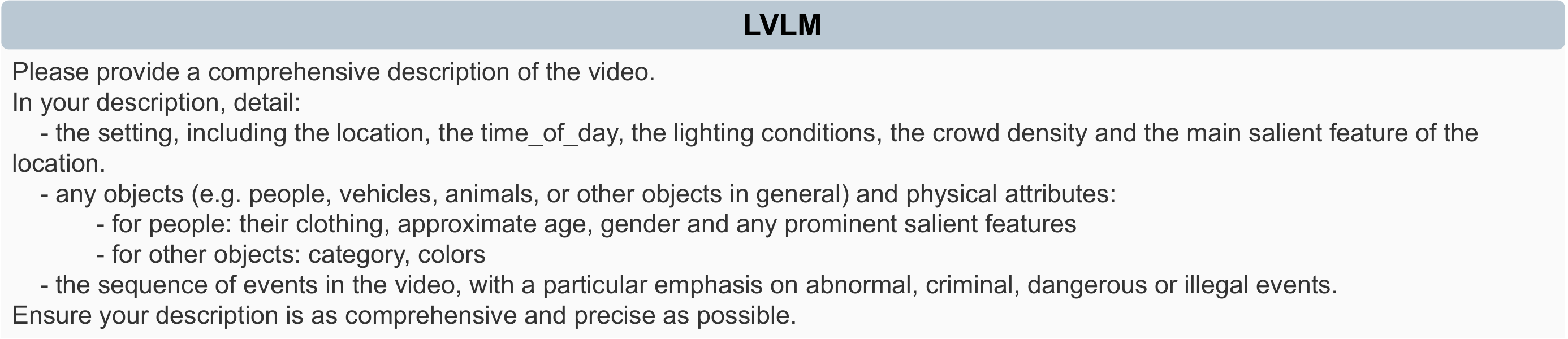}
    \caption{\textbf{Prompt for LVLM evaluation in \benchname.} We use a unified prompt across all five open-source LVLMs evaluated. This prompt is designed to extract maximum performance in our benchmark, deliberately encouraging the models to provide a report that addresses all aspects of \benchname~evaluation.}
    \label{fig:lvlm_prompt}
\end{figure*}

\begin{figure*}
    \centering
    \includegraphics[width=\linewidth]{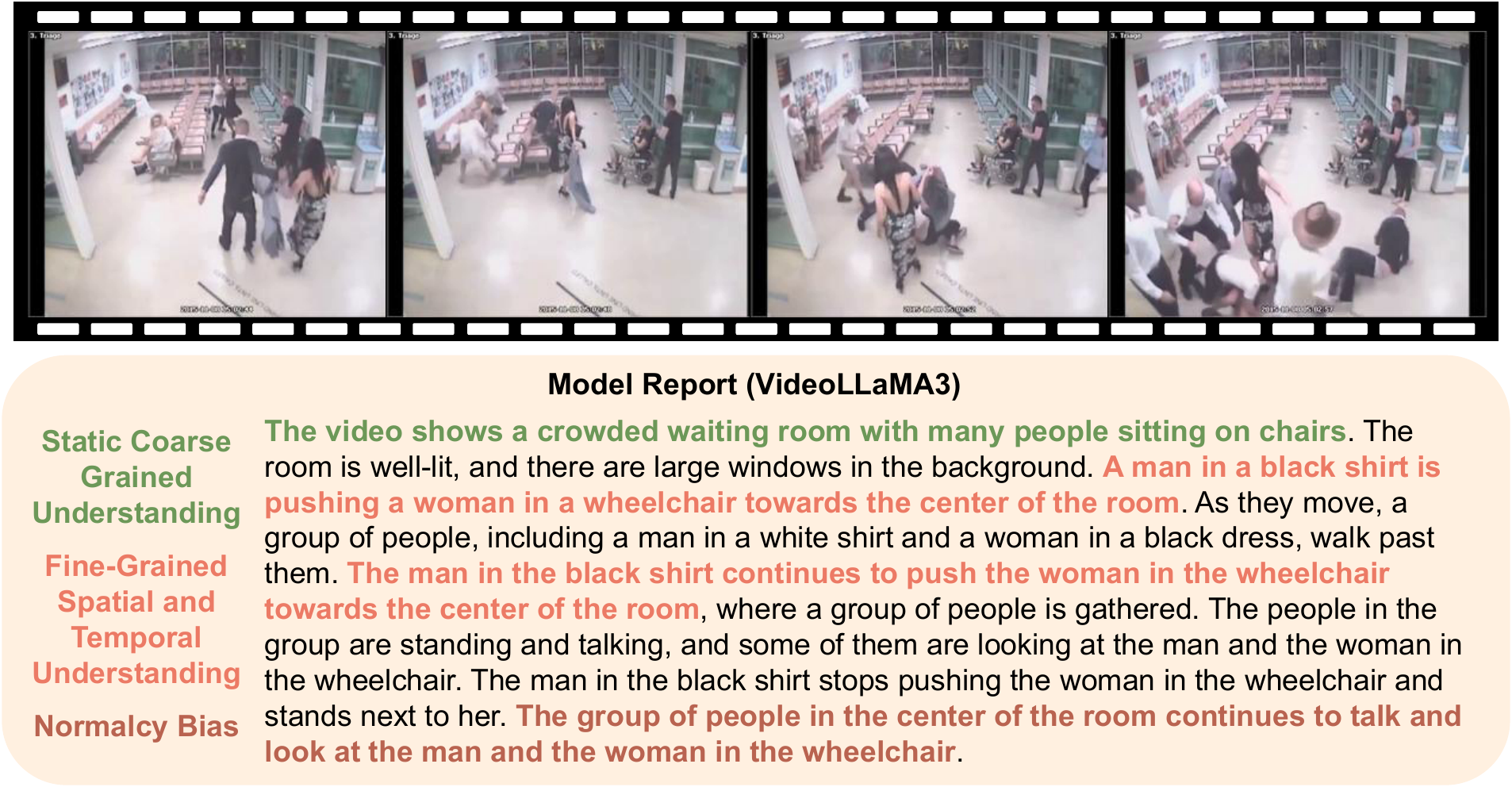}
    \caption{\textbf{Qualitative example of a weak report.} In this example, VideoLLaMA3~\cite{ref_videollama3} provides an answer that exhibits the patterns identified in the results discussion. While the model can understand the general context of the video (i.e., a "crowded waiting room with many people sitting on chairs"), it struggles to understand specific details, conflating the man in the wheelchair, with one of the women, and hallucinating normal actions such as "talking" and "pushing the woman in the wheelchair". Critically, it completely fails to report any fight or physical altercation between the man in suit and the man in white shirt, as well as the involvement of security officers and the man in a brown cowboy hat.}
    \label{fig:qualitative_1}
\end{figure*}

\begin{figure*}
    \centering
    \includegraphics[width=\linewidth]{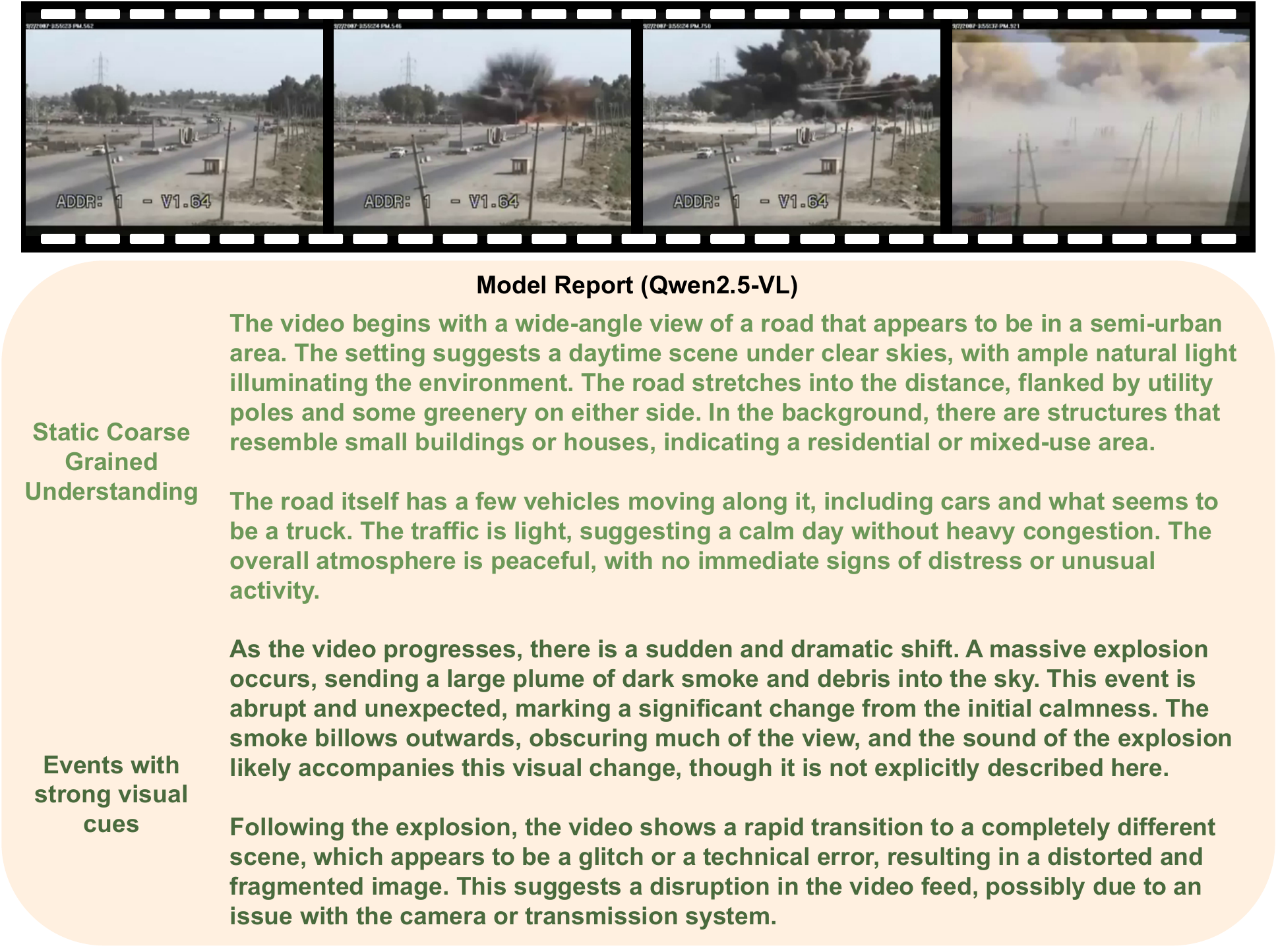}
    \caption{\textbf{Qualitative example of a performant report.} Despite the low accuracy of Qwen2.5-VL~\cite{ref_qwen25vl}, the model can successfully identify the overall scene of the video and accurately describe the explosion, the dark smoke, the debris and even the camera glitch. This example further validates our hypothesis that LVLMs can accurately report more static and coarse information, or events with strong visual cues.}
    \label{fig:qualitative_2}
\end{figure*}

\begin{figure*}
    \centering
    \includegraphics[width=\linewidth]{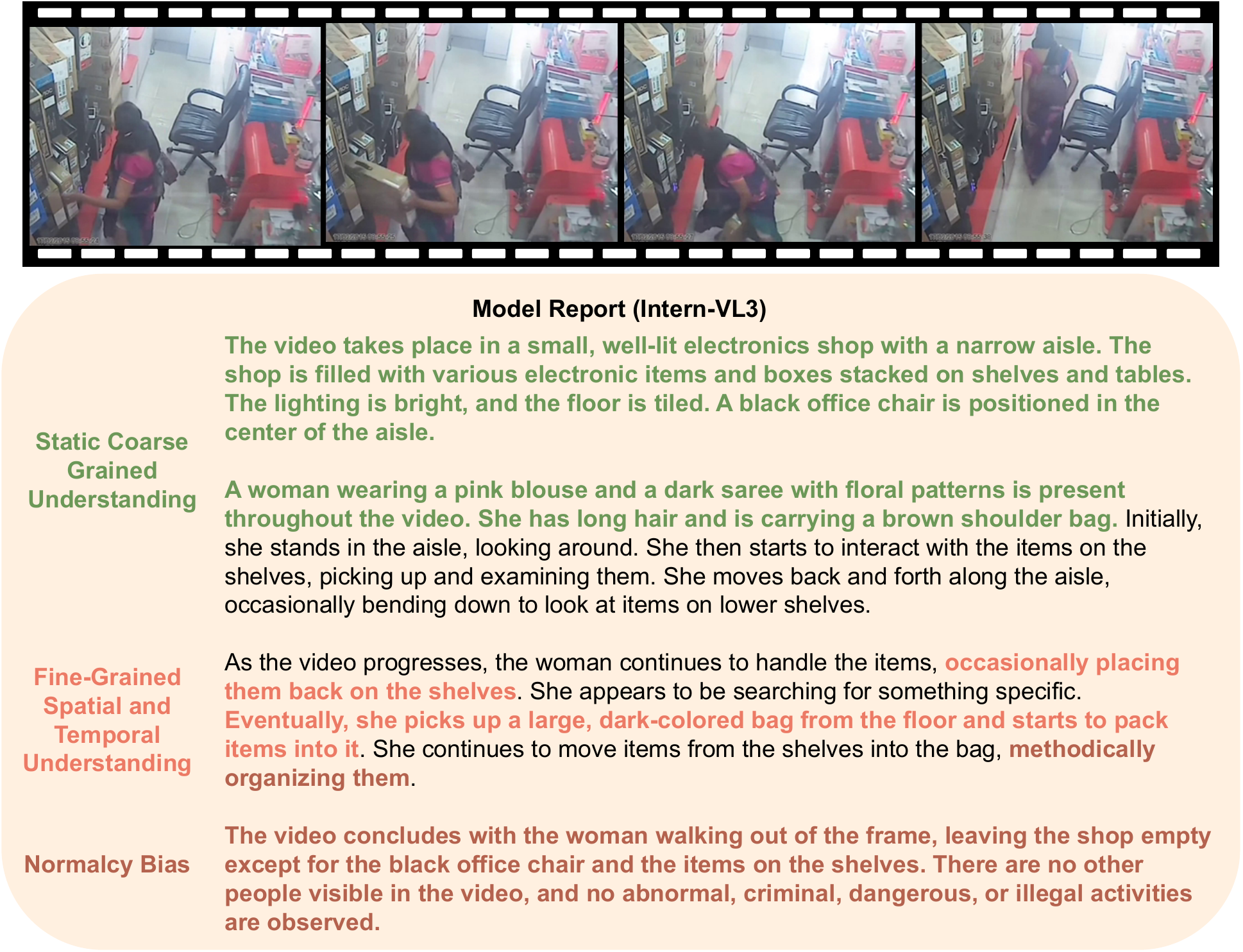}
    \caption{\textbf{Qualitative example of a mixed report.} Despite being the strongest baseline, InternVL3 still struggles with very sudden and discrete events. In this case, while the model can accurately report the context of the scene, where a woman is looking at items in an electronics shop; and the physical attributes of the woman, such as the long hair. However, it conflates the skirt of the woman with a dark-colored bag and the act of hiding the items in her purse with methodically organizing items into the bag, revealing difficulty in understanding details and a strong hallucination bias towards normal situations.}
    \label{fig:qualitative_3}
\end{figure*}

In this supplementary material, we provide additional details regarding the prompts utilized in \benchname~(Figures \ref{fig:prompt_judge}, \ref{fig:prompt_annotation_stage1}, \ref{fig:prompt_annotation_stage2} and \ref{fig:lvlm_prompt}); the annotation structure (Figure \ref{fig:annotation_example}); the type of abnormal events that are expected from each high level video category; and several qualitative examples of LVLM answers that demonstrate success and failure cases (Figures \ref{fig:qualitative_1}, \ref{fig:qualitative_2} and \ref{fig:qualitative_3}), supporting our result discussion. The \datasetname~annotations are publicly available at \url{https://huggingface.co/datasets/joao-cardeira/FineW3}.  

\section{Common Events per Video Category}

\subsection{Abuse} Contains events that involve bad, cruel or violent behavior against children, old people, animals, and women. Also includes police brutality.

\subsection{Burglary} Contains events that show people (thieves) entering into a building or house with the intention to commit theft. 

\subsection{Robbery} Contains events showing thieves taking money unlawfully by force or threat of force. Often involves using a weapon to impose fear.

\subsection{Stealing} Contains events showing people taking property or money without permission, usually on the public street. Typical examples would be stealing a vehicle.

\subsection{Shooting} Contains events showing act of shooting someone with a gun.

\subsection{Shoplifting} Contains events showing people stealing items from a shop while posing as a shopper.

\subsection{Assault} Contains events showing a sudden or violent physical attack on someone. Unlike Fighting, the person who is assaulted does not fight back and the attack is sudden and unexpected.

\subsection{Fighting} Contains events displaying two or more people involved in physical altercations, where they attack each-other.

\subsection{Arson} Contains events showing people deliberately setting fire to property, such as vehicles or homes.

\subsection{Explosion} Contains events showing destructive event of something unintentionally exploding, such as a vehicle or a factory.

\subsection{Arrest} Contains events showing police arresting individuals. These are legitimate arrests, in contrast with police brutality, considered in \textit{Abuse}. 

\subsection{Road Accident} Contains events showing traffic accidents involving vehicles, pedestrians or cyclists. 

\subsection{Vandalism} Contains events involving deliberate destruction of or damage to public or private property, including property damage, such as graffiti and defacement directed towards any property without permission of the owner.

\subsection{Normal} Contains events where no crime occurred. These videos include both indoor (such as a shopping mall) and outdoor scenes as well as day and night-time scenes.


\clearpage


\end{document}